\documentclass{llncs}
\usepackage{times}
\usepackage{epsfig}
\usepackage{graphicx}
\usepackage{amsmath}
\usepackage{amssymb}
\usepackage{algorithm}
\usepackage{algpseudocode}

\usepackage[table,xcdraw]{xcolor}
\usepackage{ragged2e}
\usepackage{multirow}
\usepackage{diagbox}
\usepackage{afterpage}
\usepackage{lipsum}

\newcommand\blfootnote[1]{%
  \begingroup
  \renewcommand\thefootnote{}\footnote{#1}%
  \addtocounter{footnote}{-1}%
  \endgroup
}

\newcommand{\argmin}{\operatornamewithlimits{argmin}}
\newcommand{\argmax}{\operatornamewithlimits{argmax}}

\title{Deformable Registration through Learning of Context-Specific Metric Aggregation}
\author{*Enzo Ferrante$^{1,2}$, *Puneet K Dokania$^{1,4}$, Rafael Marini$^{1,3}$, Nikos Paragios$^{1,3}$}
\institute{$^1$ Center for Visual Computing, CentraleSupelec, INRIA, Universite Paris-Saclay.\\
$^2$ Biomedical Image Analysis (BioMedIA) Group, Imperial College London \\
$^3$ TheraPanacea, $^4$ University of Oxford}

\begin{document}
\maketitle
\vspace{-8mm}
\begin{abstract}
\blfootnote{* indicates equal contribution.}
We propose a novel weakly supervised discriminative algorithm for learning context specific registration metrics as a linear combination of conventional similarity measures. Conventional metrics have been extensively used over the past two decades and therefore both their strengths and limitations are known. The challenge is to find the optimal relative weighting (or parameters) of different metrics forming the similarity measure of the registration algorithm. Hand-tuning these parameters would result in sub optimal solutions and quickly become infeasible as the number of metrics increases. Furthermore, such hand-crafted combination can only happen at global scale (entire volume) and therefore will not be able to account for the different tissue properties.
We propose a learning algorithm for estimating these parameters locally, conditioned to the data semantic classes. The objective function of our formulation is a special case of non-convex function, difference of convex function, which we optimize using the concave convex procedure. As a proof of concept, we show the impact of our approach on three challenging datasets for different anatomical structures and modalities.
\end{abstract}
\section{Introduction}
\vspace{-3.5mm}
Deformable image registration is a highly challenging problem frequently encountered in medical image analysis.
It involves the definition of a similarity criterion (data term) that, once endowed with a deformation model and a smoothness constraint, determines the optimal transformation to align two given images. We adopt a popular graphical model framework~\cite{GlockerDrop_MIA08} to cast deformable registration as a discrete inference problem. The definition of the data term is among the most critical components of the registration process. It refers to a function that measures the (dis)similarity between images such as mutual information ({\sc mi}) or sum of absolute differences ({\sc sad}). Metric learning in the context of image registration~\cite{Bronstein2010,Lee2009,Michel2011,Simonovsky2016,Zagoruyko2015} is an alternative that aims to determine the most efficient means of image comparison (similarity measure) from labeled visual correspondences. Our approach can be considered as a specific case of metric learning where the idea is to efficiently combine the well studied mono/multi-modal metrics depending on the local context. We aim to learn the relative weighting from a given training dataset using a learning framework conditioned on prior semantic knowledge. We propose a novel {\em weakly supervised} discriminative learning framework, based structured support vector machines ({\sc ssvm})~\cite{Taskar03M3N,Tsochantaridis04SSVM} and its extension to latent models {\sc lssvm}~\cite{Yu_LatentSSVM_09ICML}, to learn the relative weights of context specific metric aggregations.

\noindent \textbf{Metric learning.} Various metric learning methods have been proposed in the context of image registration.
Lee et al. \cite{Lee2009} introduced a multi-modal registration algorithm where the similarity measure is learned such that the target and the correctly deformed source image receive high similarity scores. The training data consisted of pre-aligned images and the learning is performed at the patch level with an assumption that the similarity measure decompose over the patches. \cite{Bronstein2010,Michel2011} proposed the use of sensitive hashing to learn a multi-modal metric. Similar to \cite{Lee2009}, they adopted a patch-wise approach. The dataset consisted of pairs of perfectly aligned images and a collection of positive/negative pairs of patches. Another patch-based alternative was presented by \cite{Toga2008} where the training set consisted of non-aligned images with manually annotated patch pairs (landmarks). 
More recently, approaches based on convolutional neural networks started to gain popularity. Zagoruyko et al. \cite{Zagoruyko2015} discussed CNN architectures to learn patch based similarity measures. One of them was then adopted in \cite{Simonovsky2016} to perform image registration. These methods require ground truth data in the form of correspondences (patches, landmarks or dense deformation fields), which is extremely difficult to obtain in real clinical data. Instead, our method is only based on segmentation masks.

\noindent \textbf{Metric aggregation.} In contrast to the above approaches, our method aggregates standard metrics using contextual information. \cite{Cifor2012} showed, in fact, that using a multichannel registration method where a set of features is globally considered instead of a single similarity measure, produced robust registration compared to using individual features. However, they did not discuss how these features can be weighted. Following this, \cite{Cifor2013a} proposed to estimate different deformation fields from each feature independently, and then compose them into final diffeomorphic transformation. Such strategy produces multiple deformation models (equal to number of metrics) which might be locally inconsistent. Thus, their combination may not be anatomically meaningful. Our method is most similar to Tang et al. \cite{Tang2012}, which generates a vector weight map that determines, at each spatial location, the relative importance of each constituent of the overall metric. However, the proposed learning strategy still requires ground truth data in the form of correspondences (pre-registered images) which is not necesseary in our case.

\noindent \textbf{Contribution.} We tackle the scenario where the ground truth deformations are not known a priori. We consider these deformation fields as latent variables, and devise an algorithm within the {\sc lssvm} framework~\cite{Yu_LatentSSVM_09ICML}. We model the latent variable imputation problem as the deformable registration problem with additional constraints. In the end, we incorporate the learned aggregated metrics in a context-specific registration framework, where different weights are used depending on the structures being registered.
\vspace{-8mm}
\section{The Deformable Registration Problem}
\vspace{-3mm}
\label{sec:regProblem}
Let us assume a source three dimensional $(3D)$ image $I$, a source $3D$ segmentation mask $S^I$ and a target $3D$ image $J$. The segmentation mask is formed by labels $s_k \in \mathcal{C}$, where $ \mathcal{C}$ is the set of classes. We focus on the 3D to 3D deformable registration problem. 
Let us also adopt without loss of generality a graphical model~\cite{GlockerDrop_MIA08,Paragios2016} for the deformable registration problem. A deformation field is sparsely represented by a regular grid graph $G = (V, E)$, where $V$ is the set of nodes and $E$ is the set of edges. Each node $i \in V$ corresponds to a control point $p_i$. Each control point $p_i$ is allowed to move in the $3D$ space, therefore, can be assigned a label $\boldsymbol{d_i}$ from the set of $3D$ displacement vectors $\mathcal{L}$. Notice that each $3D$ displacement vector is a tuple defined as $\boldsymbol{d_i} = \{dx_i, dy_i, dz_i\}$, where $dx$, $dy$, and $dz$ are the displacements in the $x$, $y$, and $z$ directions, respectively. The deformation (labeling of the graph $G$) denoted as $D \in \mathcal{L}^{|V|}$ is associated to a set of nodes $V$, where each node is assigned a displacement vector $\boldsymbol{d_i}$ from the set $\mathcal{L}$. The new control point obtained when the displacement $\boldsymbol{d_i}$ is applied to the original control point $p_i$ is denoted as $\bar{p}_i$. Let us define a patch $\bar{\Omega}_i^I$ on the source image $I$ centered at the displaced control point $\bar{p}_i$. Similarly, we define $\Omega_i^J$ as the patch on the target image $J$ centered at the original control point $p_i$, and $\bar{\Omega}_i^{S^I}$ as the patch on the input segmentation mask centered at the displaced control point $\bar{p}_i$. Using the above notations, we define the unary feature vector corresponding to the $i^{th}$ node for a given displacement vector $\boldsymbol{d_i}$ as $\mathcal{U}_i(\boldsymbol{d_i}, I, J) = (u_1(\bar{\Omega}_i^I, \Omega_i^J), \cdots,  u_n(\bar{\Omega}^I, \Omega_i^J)) \in \mathbb{R}^n$, where $n$ is the number of metrics (or similarity measures) and $u_j(\bar{\Omega}_i^I, \Omega_i^J)$ is the unary feature corresponding to the $j^{th}$ metric on the patches $\bar{\Omega}_i^I$ and $\Omega_i^J$. 
In case of single metric, we define $n=1$. Therefore, given a weight matrix $W \in \mathbb{R}^{n \times |\mathcal{C}|}$, where $W(i,j)$ denote the weight of the $i^{th}$ metric corresponding to the class $j$, the unary potential of the $i^{th}$ node for a given displacement vector $\boldsymbol{d_i}$ is computed as:
\vspace{-1mm}
\begin{align}
\label{eq:unaryPotential}
\bar{\mathcal{U}}_i(\boldsymbol{d_i}, I, J, S^I; W) = {\bf w}(\bar{c})^\top \mathcal{U}_i(\boldsymbol{d_i}, I, J) \in \mathbb{R}.
\end{align}
where, ${\bf w}(\bar{c}) \in \mathbb{R}^n$ is the $\bar{c}^{th}$ column of the weight matrix $W$ and $\bar{c}$ is the most dominant class in the patch on the source segmentation mask $\bar{\Omega}_i^{S^I}$ obtained as $\bar{c}=\argmax_{c \in \mathcal{C}} f(\bar{\Omega}_i^{S^I}, c)$, with $f(\bar{\Omega}_i^{S^I}, c)$ being the number of voxels of class $c$ in the patch $\bar{\Omega}_i^{S^I}$. Other criterion could be used to find the dominant class. The pairwise clique potential between the control points $p_i$ and $p_j$ is defined as $\mathcal{V}(\boldsymbol{d_i}, \boldsymbol{d_j})$, where $\mathcal{V}(.,.)$ is the $L_1$ norm between the two input arguments. Thus, the multi-class energy function is: 
\vspace{-2mm}
\begin{equation}
\label{eq:energyFunctionFull}
\mathcal{E}(I,J,S^I,D; W) = \sum_{i \in V} \bar{\mathcal{U}}_i(\boldsymbol{d_i}, I, J, S^I; W) + \sum_{(i,j) \in E} \mathcal{V}(\boldsymbol{d_i}, \boldsymbol{d_j})
\end{equation}
Then, the optimal deformation is obtained as $\hat{D} = \argmin_{D \in \mathcal{L}^{|V|}} \mathcal{E}(I,J,S^I,D; W).$
This problem is {\sc np-hard} in general. Similar to~\cite{GlockerDrop_MIA08}, we adopt a pyramidal approach to solve the problem efficiently. 
We use FastPD~\cite{KomodakisFastPD_CVPR07} for the inference at every level of the pyramid. Notice that the energy function~(\ref{eq:energyFunctionFull}) is defined over the nodes of the sparse graph $G$.
Once we obtain the optimal deformation $\hat{D}$, we estimate the dense deformation field using a free form deformation (FFD) model~\cite{RueckertFFD_TMI99} in order to warp the input image. 
\vspace{-10mm}
\section{Learning the Parameters}
\vspace{-4mm}
\label{sec:learningParameters}
Knowing the weight matrix $W$ a priori is non-trivial and hand tuning it quickly becomes infeasible as the number of metrics and classes increases. We propose an algorithm to learn $W$ conditioned on the semantic labels assuming that in the training phase semantic masks are available for the source and the target images. Instead of learning the complete weight matrix at once, we learn the weights (or parameters) for each class $c \in \mathcal{C}$ individually. Now onwards, the weight vector ${\bf w}_c$ denotes a particular column of the weight matrix $W$, representing the weights corresponding to the $c^{th}$ class. 

\noindent \textbf{Training Data.} Consider a dataset $\mathcal{D} = \{( {\bf x}_i, {\bf y}_i )\}_{i=1,\cdots,N}$, where ${\bf x}_i = (I_i, J_i)$, $I_i$ is the source image and $J_i$ is the target. Similarly, ${\bf y}_i = (S^I_i, S^J_i)$, where $S^I_i$ and $S^J_i$ are the segmentation masks for the source and target images. The size of each segmentation mask is the same as that of the corresponding images. As stated earlier, the segmentation mask is formed by the elements (or voxels) $s_k \in \mathcal{C}$, where $ \mathcal{C}$ is the set of classes. 

\vspace{0.5mm}
\noindent \textbf{Loss Function.}
The loss function $\Delta(S^I, S^J) \in \mathbb{R}_{\geq 0}$ evaluates the similarity between the segmentation masks $S^I$ and $S^J$. Higher $\Delta(., .)$ implies higher dissimilarity. We use a dice based loss function as this is our evaluation criteria:
\begin{equation} 
\label{eq:Loss}
\Delta(S^I, S^J)  = 1 - DICE(S^I, S^J) = 1 - (2 \sum_{i \in V} \frac{|\phi(p_i^I) \cap \phi(p_i^J)|}{|\phi(p_i^I)| + |\phi(p_i^J)|}),
\end{equation}
where, $\phi(p_i^I)$ and $\phi(p_i^J)$ are the patches at the control point $p_i$ on the segmentation masks $S^I$ and $S^J$, respectively, and $|.|$ represents cardinality. This approximation makes the dice decomposable over the nodes of $G$ enabling a very efficient training.

\noindent \textbf{Joint Feature Map.} Given ${\bf w}_c$ for the $c$-th class, the deformation $D$ and input ${\bf x}$, the multi-class function~(\ref{eq:energyFunctionFull}) can be trivially converted into class-based energy function as:
\vspace{-1mm}
\begin{equation}
\label{eq:energyFunctionPerClass}
	\mathcal{E}_c({\bf x},D; {\bf w}) = {\bf w}_c^\top \sum_{i \in V} \mathcal{U}_i(\boldsymbol{d_i}, {\bf x}) + w_p \sum_{(i,j) \in E} \mathcal{V}(\boldsymbol{d_i}, \boldsymbol{d_j}),
\end{equation}
where $w_p \in \mathbb{R}_{\geq 0}$ is the parameter for the pairwise term. The final parameter vector ${\bf w} \in \mathbb{R}^{n+1}$ is the concatenation of ${\bf w}_c$ and $w_p$. Thus, the function~(\ref{eq:energyFunctionPerClass}) can be written as:
\vspace{-1mm}
\begin{equation}
\label{eq:energyFunctionPsi}
	\mathcal{E}_c({\bf x},D; {\bf w}) = {\bf w}^\top \Psi({\bf x}, D),
\end{equation}
where $\Psi({\bf x}, D) \in \mathbb{R}^{n+1}$ is the joint feature map defined as:
\vspace{-1mm}
\begin{equation}
\label{eq:Psi}
	\Psi ({\bf x}, D) = 
	\begin{pmatrix} \sum_{i \in V} \mathcal{U}^1_i(\boldsymbol{d_i}, {\bf x}) \\ \vdots \\ \sum_{i \in V} \mathcal{U}^{n}_i(\boldsymbol{d_i}, {\bf x}) \\ \sum_{(i,j) \in E} \mathcal{V}(\boldsymbol{d_i}, \boldsymbol{d_j}) \\ \end{pmatrix}
\end{equation}\\
Notice that the energy function~(\ref{eq:energyFunctionPerClass}) does not depend on the source segmentation mask $S^I$. The only use of the source segmentation mask in the energy function~(\ref{eq:energyFunctionFull}) is to obtain the dominant class 
which in this case is not required. However, we will shortly see that the source segmentation mask $S^I$ plays a crucial role in the learning algorithm.
\begin{algorithm}[tb]
\caption{The {\sc cccp} Algorithm.}
\label{algo:hierPn}
\begin{algorithmic}[1]
\State  $\mathcal{D}$, ${\bf w}_0$, $C$, $\alpha$, $\eta$, the tolerance $\epsilon$.
\State $t=0$, ${\bf w}_t = {\bf w}_0$.
\Repeat
\State For a given ${\bf w}_t$, impute latent variables $\hat{D}_i$ for each sample by solving~(\ref{eq:imputeLatentRelaxed}).
\State Update parameters ${\bf w}_{t+1}$ by optimizing the convex optimization problem~(\ref{eq:learningObjectiveSSVM}). 
\State $t=t+1$
\Until{ The objective function of the problem~(\ref{eq:learningObjective}) does not decrease more than $\epsilon$}.
\end{algorithmic}
\label{algo:cccp}
\end{algorithm}
\noindent \textbf{Latent Variables.} Ideally, the dataset $\mathcal{D}$ must contain the ground truth deformations $D$ corresponding to the source image $I$ in order to compute the energy term defined in the equation~(\ref{eq:energyFunctionPerClass}). Since annotating the dataset with the ground truth deformation is non-trivial, we use them as the latent variables in our algorithm. 

\noindent \textbf{The Objective Function.}
Given $\mathcal{D}$, we learn the parameter ${\bf w}$ such that minimizing the energy function~(\ref{eq:energyFunctionPerClass}) leads to a deformation field which when applied to the source segmentation mask gives minimum loss with respect to the target segmentation mask. We denote $g(S,D)$ as the deformed segmentation when the dense deformation field obtained from $D$ is applied to the segmentation mask $S$. 
Similarly to the latent {\sc ssvm}~\cite{Yu_LatentSSVM_09ICML}, we optimize a regularized upper bound on the loss:
\vspace{-2mm}
\begin{eqnarray}
\label{eq:learningObjective}
&&\min_{\bf w, \{\xi_i\}} \frac{1}{2}||{\bf w}||^2 + \alpha ||{\bf w} - {\bf w}_0 ||^2 + \frac{C}{N} \sum_i \xi_i , \nonumber \\
&& s.t. \min_{D,  \Delta(g(S_i^I, D), S_i^J) =0} {\bf w}^\top \Psi({\bf x}_i, D) \leq {\bf w}^\top \Psi({\bf x}_i, \bar{D}) - \Delta(g(S_i^I, \bar{D}), S_i^J) + \xi_i, \nonumber \\
&& \forall \bar{D}, w_p \geq 0, \xi_i \geq 0, \forall i.
\end{eqnarray}
where, $\bar{D} = \argmin_{D} \mathcal{E}({\bf x}_i,D; {\bf w})$. The above objective function minimizes an upper bound on the given loss, called slack ($\xi_i$). The effect of the regularization term is controlled by the hyper-parameter $C$. The second term is the proximity term to ensure that the learned ${\bf w}$ is close to the initialization ${\bf w}_0$. The effect of the proximity term is controlled by the hyperparameter $\alpha$. Intuitively, for a given input-output pair, the constraints of the above objective function enforce that the energy corresponding to the best possible deformation field, in terms of both energy and loss (in order to be semantically meaningful), must always be less than or equal to the energy corresponding to any other deformation field with a margin proportional to the loss and some non negative slack. 

\vspace{0.5mm}
\noindent \textbf{The Learning Algorithm.} The objective function~(\ref{eq:learningObjective}) turns out to be a special case of non-convex functions (difference of convex), thus can be locally optimized using the well known {\sc cccp} algorithm~\cite{Yuille03CCCP}. The {\sc cccp} algorithm consist of three steps -- (1) upperbounding the concave part at a given ${\bf w}$, which leads to an affine function in ${\bf w}$; (2) optimizing the resultant convex function (sum of convex and affine functions is convex); (3) repeating the above steps until the objective can not be further decreased beyond a given tolerance of $\epsilon$. The complete {\sc cccp} algorithm for the optimization of~(\ref{eq:learningObjective}) is shown Algorithm~\ref{algo:cccp}. The first step of upperbounding the concave functions (Line 4) is the same as the latent imputation step, which we call the {\em segmentation consistent registration} problem. The second step is the optimization of the resultant convex problem (Line 5), which is the optimization of the {\sc ssvm} for which we use the  well known cutting plane algorithm~\cite{Joachims09_CuttingPlane}. In what follows, we discuss these steps in detail.

\noindent \textbf{Segmentation Consistent Registration.} This step involves generating the best possible ground truth deformation field (unknown a priori) at a given ${\bf w}$, known as the latent imputation step. Since we optimize the dice loss, we formulate this step as an inference problem with additional constraints to ensure that the imputed deformation warps the image minimizing the loss between the deformed source and the target. Mathematically, for a given parameter vector ${\bf w}$, the latent deformation is imputed by solving:
\vspace{-2mm}
\begin{align}
\label{eq:imputeLatent}
\hat{D}_i = \argmin_{D \in \mathcal{L}^{|V|},  \Delta(g(S_i^I, D), S_i^J) =0} {\bf w}^\top \Psi({\bf x}_i, D).
\end{align}
We relax the above problem as it is difficult and may not have a unique solution: 
\vspace{-2mm}
\begin{align}
\label{eq:imputeLatentRelaxed}
\hat{D}_i = \argmin_{D \in \mathcal{L}^{|V|}} \Big( {\bf w}^\top \Psi({\bf x}_i, D) + \eta \Delta(g(S_i^I, D), S_i^J) \Big),
\end{align}
where, $\eta \geq 0$ controls the relaxation trade-off. Since the loss function used is decomposable, the above problem can be optimized using FastPD inference for the deformable registration with trivial modifications on the unary potentials.

\vspace{-0.5mm}
\noindent \textbf{Parameters update.} Given the imputed latent variables, the resultant objective is:
\vspace{-2mm}
\begin{eqnarray}
\label{eq:learningObjectiveSSVM}
&&\min_{\bf w, \{\xi_i\}} \frac{1}{2}||{\bf w}||^2 + \alpha ||{\bf w} - {\bf w}_0 ||^2 + \frac{C}{N} \sum_i \xi_i , \nonumber \\
&&s.t. {\bf w}^\top \Psi({\bf x}_i, \hat{D}_i) \leq \hspace{-2pt} {\bf w}^\top \Psi({\bf x}_i, \bar{D}) - \Delta(g(S_i^I, \bar{D}), S_i^J) + \xi_i, \forall \bar{D}, w_p, \xi_i \geq 0, \forall i.
\end{eqnarray}
where, $\hat{D}_i$ is the latent deformation field imputed by solving the problem~(\ref{eq:imputeLatentRelaxed}). Intuitively, the above objective function tries to learn the parameters ${\bf w}$ such that the energy corresponding to the imputed deformation field is always less than the energy for any other deformation field with a margin proportional to the loss function. The above objective function has exponential number of constraints, one for each possible deformation field $\bar{D} \in \mathcal{L}^{|V|}$. In order to alleviate this problem we use cutting plane algorithm~\cite{Joachims09_CuttingPlane}. Briefly, for a given ${\bf w}$, each deformation field $\bar{D}$ gives a slack. Instead of minimizing all the slacks for a particular sample at once, we find the deformation field that leads to the maximum value of the slack and store this in a set known as the working set. This is known as {\em finding the most violated constraint}. Thus, instead of using exponentially many constraints, the algorithm uses the constraints stored in the working set and this process is repeated until no constraints can be further added. Rearranging the terms in the constraints of the objective function~(\ref{eq:learningObjectiveSSVM}) and ignoring the constant term ${\bf w}^\top \Psi({\bf x}_i, \hat{D}_i)$, the most violated constraint can be obtained by solving:
\vspace{-2mm}
\begin{align}
\label{eq:mostViolated}
\bar{D}_i =  \argmin_{D \in \mathcal{L}^{|V|}} \Big( {\bf w}^\top \Psi({\bf x}_i, \bar{D}) - \Delta(g(S_i^I, \bar{D}), S_i^J ) \Big).
\end{align}
Since the loss is decomposable, this problem can be solved using FastPD inference for the deformable registration with trivial modifications on the unary terms.

\vspace{0.5mm}
\noindent \textbf{Prediction.} Once we obtain the learned parameters ${\bf w}_c$ for each class  $c \in \mathcal{C}$ using the Algorithm~\ref{algo:cccp}, we form the matrix $W$ where each column of the matrix represents the learned parameter for a specific class. This $W$ is then used to solve the registration problem (equation~(\ref{eq:energyFunctionFull})) using the approximate inference discussed in Section~\ref{sec:regProblem}.
\vspace{-4mm}
\section{Results and discussion}
\begin{figure*}[t]
  \centering
  \includegraphics[width=1\textwidth]{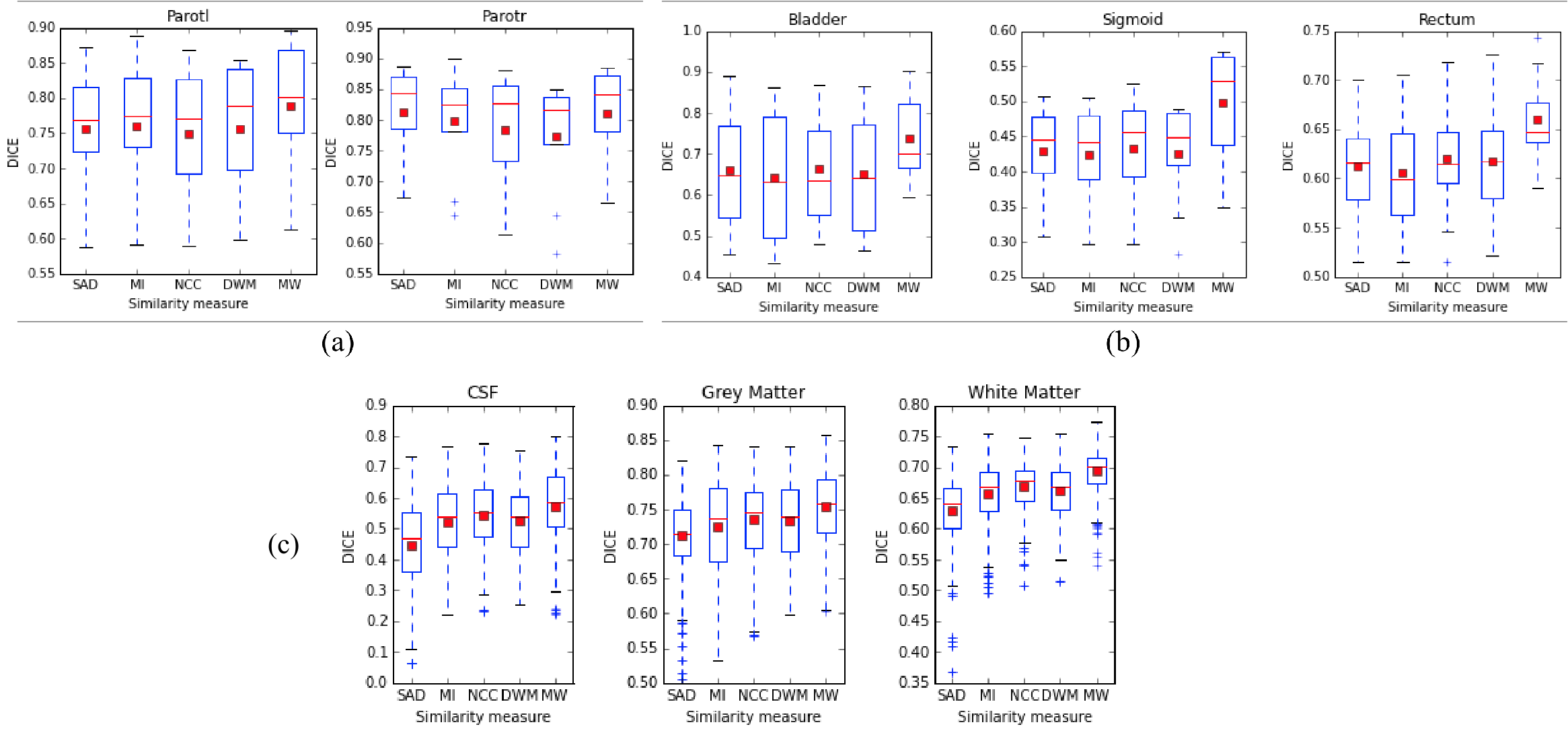}
  \caption{Results for RT parotids (a), RT abdominal (b) and IBSR (c) datasets. We show dice between the deformed source and the target segmentation masks after registration, for the single-metric registration ({\sc sad}, {\sc mi}, {\sc ncc}, {\sc dwt}) and the learned multi-metric registration ({\sc mw}). In (a), `Parotl' and `Parotr' are the left and the right parotids. In (b), `Bladder', `Sigmoid', and `Rectum' are the three organs in the dataset. In (c), annotations correspond to Cerebrospinal fluid (CSF),grey (GM) and white (WM) matter. The red square is the mean and the red bar is median.
  } 
  \label{fig:allResults}
\end{figure*}

\vspace{-4mm}
\label{sec:experiments}
As a proof of concept, we evaluate the effect of the aggregated metric on three different medical datasets -- (1) RT Parotids, (2) RT Abdominal, and a downsampled version of (3) IBSR \cite{ibsr}, involving several anatomical structures, different image modalities, and inter/intra patient images
We used four different metrics: (1) sum of absolute differences ({\sc sad}), (2) mutual information ({\sc mi}), (3) normalized cross correlation ({\sc ncc}), and (4) discrete wavelet coefficients ({\sc dwt}). 
The datasets consists of 8 CT (RT Parotids, head images of $56\times62\times53$ voxels), 5 CT (RT Abdominal, abdominal images of $90\times60\times80$ voxels) and 18 MRI images (a downsampled version of IBSR dataset, including brain images $64\times64\times64$ voxels). We performed muli-fold cross validation in every dataset, considering pairs with different patients in training and testing. For a complete description of the datasets and the experimental setting, please refer to the supplementary material. The results on the test sets are shown in Figure~\ref{fig:allResults}. 
\begin{figure}[t!]
  \centering
  \includegraphics[width=\textwidth]{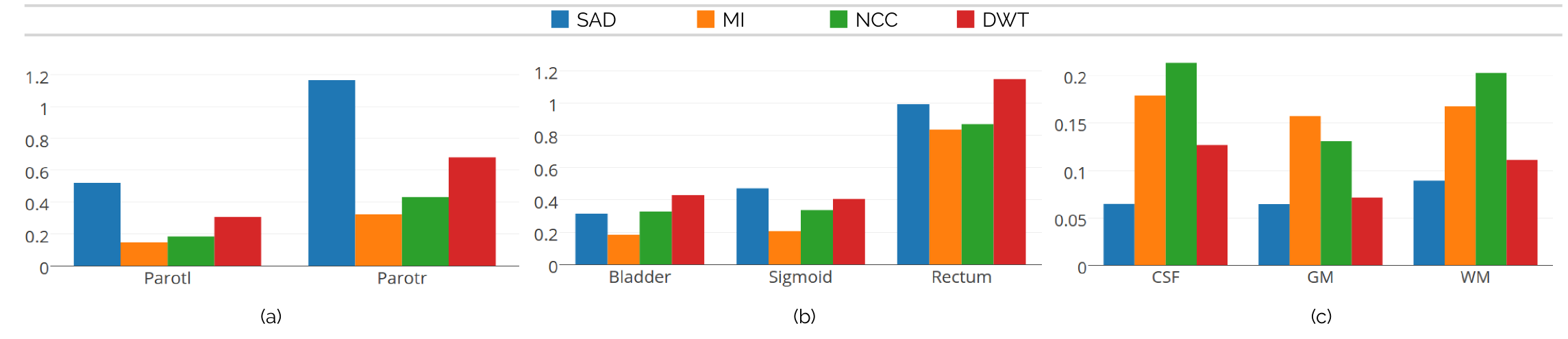}
  \caption{Example of learned weights for RT Parotids (a), RT Abdominal (b) and IBSR (c) datasets. Since the structures of interest in every dataset present different intensity distributions, different metric aggregations are learned. Note that in case of the RT Parotids, given that both parotid glands present the same intensity distribution, similar weightings are learned for both structures, with SAD dominating the other similarity measures. However, in IBSR dataset, NCC dominates in case of CSF and WM, while MI receives the higher value for gray matter (GM). } 
  \label{fig:learnedWeights}
\end{figure}
As it can be observed in Figure~\ref{fig:allResults}, the linear combination of similarity measures weighted using the learned coefficients systematically outperforms (or is as good as) single metric based registration, with max improvements of 8\% in terms of dice. 

\noindent \textbf{Discussion and conclusions.} 
We have showed that associating different similarity criteria to every anatomical region yields results superior to the classic single metric approach. In order to learn this mapping where ground truth is generally given in the form of segmentation masks, we defined deformation fields as latent variables and proposed a {\sc lssvm} based framework. The main limitation of our method is the need of segmentation masks for the source images in testing time. However, different real scenarios like radiation therapy or atlas-based segmentation methods fulfill this condition. Note that, at prediction (testing) time, the segmentation mask is used to determine the metrics weights combination per control node (finding the dominant class). The segmentation labels are not used at testing time to guide the registration process which is purely image based. In our multi-metric registration approach, segmentation masks are only required (at testing time) for the source image and used to choose the best learned metric aggregation. The idea could be further extended to unlabeled data (as it concerns the source image at testing time) where the dominant label class per control node is the output of a classification/learning method.
From a theoretical viewpoint, we showed how the three main components of LSSVM: (1) latent imputation (Eq. \ref{eq:imputeLatentRelaxed}); (2) prediction (optimizing Eq.~\ref{eq:energyFunctionFull}) and (3) finding most violated constraint (Eq.~(\ref{eq:mostViolated})), can be formulated as the exact same problem. 
The difference among these problems is the unary potentials used. This is extremely important given that further improvements in inference algorithms will directly increase the quality of the results.
As future work, the integration of alternative accuracy measures, other than dice, such as the Hausdorff distance between surfaces or real geometric distances for anatomical landmarks could further enhance the performance of the method.
\vspace{-5mm}

\bibliographystyle{splncs03}
\bibliography{dokaniaBibliography}
\newpage
\section*{Supplementary Material}

\section{Detailed experimental setting description}
In what follows, we provide a detailed description of the experimental setup used for the evaluation presented in the main paper.
For all the experiments, we used the same set of parameters for the pyramidal based inference discussed in Section~2: $2$ pyramid levels, $5$ refinement steps per pyramid level, $125$ labels, and distance between control points of $25$mm in the finer level. The running time for each registration case was around $12$ seconds. For the training, we initialized ${\bf w}_0$ with the hand tuned values (obtained using grid search for the values \{0.01, 0.1, 1, 10\} for each metric: ${\bf w}_0 = (0.1, 10, 10, 10)$, for {\sc sad}, {\sc mi}, {\sc ncc}, and {\sc dwt}, respectively.

The images used for evaluation correspond to the following datasets:
\begin{itemize}
 \item \textbf{RT Parotids.} This dataset contains $8$ CT volumes of head, obtained from 4 different patients, 2 volumes per patient. The volumes are captured in two different stages of a radiotherapy treatment in order to estimate the radiation dose. Right and left parotid glands were segmented by the specialists in every volume. The dimensions of the volumes are $56\times62\times53$ voxels with a physical spacing of $3.45$mm, $3.45$mm, and $4$mm, in x, y, and z axes, respectively. We generated $8$ pairs of source and target volumes using the given dataset. Notice that, while generating the source and target pairs, we did not mix the volumes coming from different patients. We splitted the dataset into train and test. The average results on the test set are shown in Figure 1.a from the main paper. 
 \item \textbf{RT Abdominal.} The second dataset contains $5$ CT volumes of abdomen for a particular patient captured with a time window of about 7 days during a radiotherapy treatment. Three organs have been manually segmented by the specialists: (1) sigmoid, (2) rectum, and (3) bladder. The dimensions of the volumes are $90\times60\times80$ voxels with a physical spacing of $3.67$mm, $3.67$mm, and $4$mm, in x, y, and z axes, respectively (there are small variations depending on the volume). We generated a train dataset of 6 pairs and test dataset of 4 pairs. The results on the test set are shown in Figure 1.b from the main paper.
\item \textbf{IBSR.} The third dataset (IBSR) is the well known Internet Brain Segmentation Repository dataset, which consists of $18$ brain {\sc mri} volumes, coming from different patients. Segmentations of three different brain structures are provided by the specialists: white mater (WM), gray mater (GM), and cerebrospinal fluid (CSF). We used a downsampled version of the dataset to reduce the computation cost. The dimension of the volumes are $64\times64\times64$ voxels with a physical spacing of $3.75$mm, $3.75$mm, and $3$mm in x, y, and z axes, respectively. We divided the 18 volumes in 2 folds of 9 volumes on each fold, giving total of 72 pairs per fold. We used an stochastic approach for the learning process, where we sample 10 different pairs from the training set, and we tested on the 72 pairs of the other fold. We run this experiment 3 times per fold, giving a total of 6 different experiments, with 72 testing samples and 10 training samples randomly chosen. The results on the test set are shown in Figure 1.c from the main paper.
\end{itemize}

For all the datasets, experiments were performed in two steps. First, we learned the weighting vector ${\bf w}_c$ independently for every organ $c \in \mathcal{C}$. Second, we plugged the learned weights in the multi-metric registration algorithm and we register every testing case using the method presented in Section 2. We also run experiments using single metrics (sum of absolute differences (SAD), mutual information (MI), normalized crossed correlation (NCC) and discrete wavelet transform (DWT)) with hand tuned weights obtained using a simple grid search. Results are summarized in Figure 1 from the main paper (detailed numerical values are included in Table \ref{tab:averageDiceImprovement}). As it can be observed, the trained multi-metric algorithm outperforms the single metric approaches in all the organs. 

\section{Quantitative Results}
The following table contains the numerical results corresponding to Figure 1 from the main paper.
\begin{table}[h!]
\centering
\resizebox{\columnwidth}{!}{%
\begin{tabular}{|
>{\columncolor[HTML]{EFEFEF}}c |
>{\columncolor[HTML]{EFEFEF}}c |c|c|c|c|c|c|}
\hline
\cellcolor[HTML]{C0C0C0}\textbf{Dataset}               & \cellcolor[HTML]{C0C0C0}\textbf{Organ} & \cellcolor[HTML]{C0C0C0}\textbf{SAD} & \cellcolor[HTML]{C0C0C0}\textbf{MI} & \cellcolor[HTML]{C0C0C0}\textbf{NCC} & \cellcolor[HTML]{C0C0C0}\textbf{DWT} & \cellcolor[HTML]{C0C0C0}\textbf{MW} & \cellcolor[HTML]{C0C0C0}\textbf{Average dice increment for MW} \\ \hline
\cellcolor[HTML]{EFEFEF}                               & Parotl                                 & 0,756                                & 0,760                               & 0,750                                & 0,757                                & \textbf{0,788}                               & 0,033                                                          \\ \cline{2-8} 
\multirow{-2}{*}{\cellcolor[HTML]{EFEFEF}RT Parotids}  & Parotr                                 & 0,813                                & 0,798                               & 0,783                                & 0,774                                & \textbf{0,811}                               & 0,019                                                          \\ \hline
\cellcolor[HTML]{EFEFEF}                               & Bladder                                & 0,661                                & 0,643                               & 0,662                                & 0,652                                & \textbf{0,736}                               & 0,082                                                          \\ \cline{2-8} 
\cellcolor[HTML]{EFEFEF}                               & Sigmoid                                & 0,429                                & 0,423                               & 0,432                                & 0,426                                & \textbf{0,497}                               & 0,070                                                          \\ \cline{2-8} 
\multirow{-3}{*}{\cellcolor[HTML]{EFEFEF}RT Abdominal} & Rectum                                 & 0,613                                & 0,606                               & 0,620                                & 0,617                                & \textbf{0,660}                               & 0,046                                                          \\ \hline
\cellcolor[HTML]{EFEFEF}                               & CSF                                    & 0,447                                & 0,520                               & 0,543                                & 0,527                                & \textbf{0,546}                               & 0,037                                                          \\ \cline{2-8} 
\cellcolor[HTML]{EFEFEF}                               & GM                                     & 0,712                                & 0,725                               & 0,735                                & 0,734                                & \textbf{0,761}                               & 0,035                                                          \\ \cline{2-8} 
\multirow{-3}{*}{\cellcolor[HTML]{EFEFEF}IBSR}         & WM                                     & 0,629                                & 0,658                               & 0,669                                & 0,661                                & \textbf{0,682}                               & 0,028                                                          \\ \hline
\end{tabular}
}
\caption[Average dice value per organ, for the single and multi-metric approaches, are reported for the three datasets.]{Average dice value per organ, for the single and multi-metric approaches, are reported for the three datasets. The last column indicates the average dice improvement that our proposed method provides when compared with the single metric approaches. We can observe improvements of a maximum of 8\% points in terms of dice coefficient.}
\label{tab:averageDiceImprovement}
\end{table}

\newpage

\section{Qualitative results}

Below we show visual results on three datasets used as a proof-of-concept for our proposed method, to highlight the effects of learning the weights of different metrics for the task of deformable registration.

\begin{figure}
  \centering
  \includegraphics[scale=0.6]{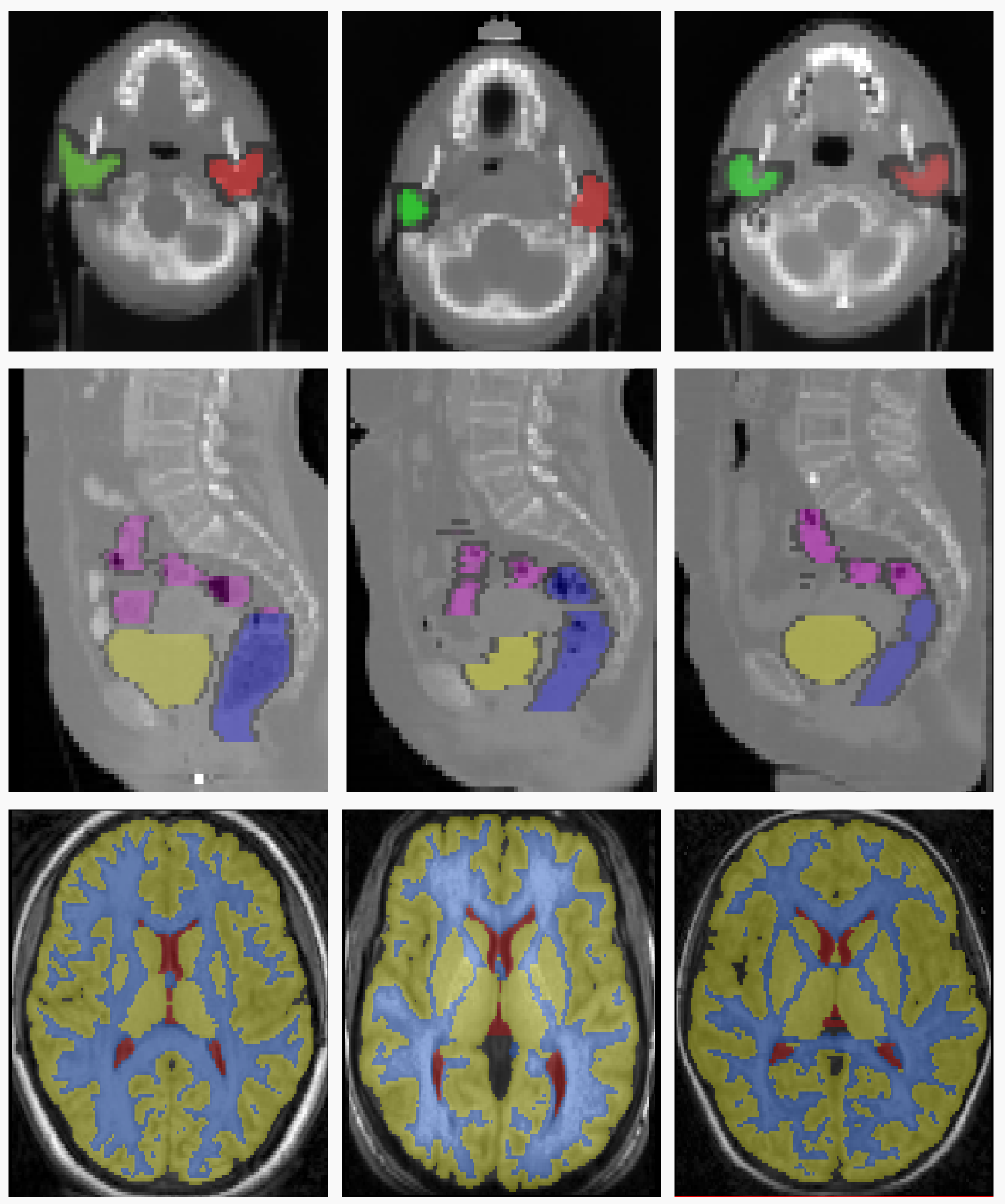}
  \caption{Sample slices from three different volumes of the RT Parotids, RT Abdominal and IBSR datasets.The top row represents the sample slices from three different volumes of the RT Parotids dataset. The middle row represents the sample slices of the RT Abdominal dataset, and the last row represents the sample slices from the IBSR dataset.} 
  \label{fig:imageExamples}
\end{figure}

\begin{figure}
  \centering
  \includegraphics[width=\textwidth]{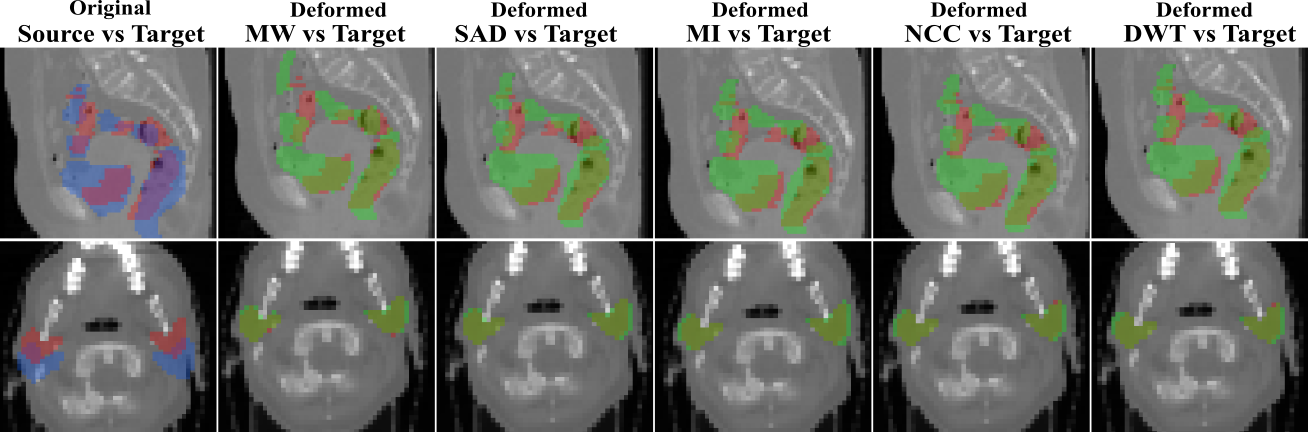}
  \caption{Overlapping of the segmentation masks in different views for one registration case from {\bf RT Abdominal} (first row) and {\bf RT Parotids} (secibd row) datasets (views are different than those shown in the main paper). The first column corresponds to the overlapping before registration between the source (in blue) and target (in red) segmentation masks of the different anatomical structures of both datasets. From second to sixth column, we observe the overlapping between the warped source (in green) and the target (in red) segmentation masks, for the multiweight algorithm (MW) and for the single metric algorithm using sum of absolute differences (SAD), mutual information (MI), normalized cross correlation (NCC) and discrete wavelet transform (DWT) as similarity measure. We observe in a qualitative way that multiweight algorithm gives a better fit between the deformed and ground truth structures than the rest of the single similarity measures, which are over segmenting most of the structures showing a poorer registration performance. This is coherent with the numerical results reported in Figures 2 and 3 from the main paper.} 
  \label{fig:resultsAustriaLebanon}
\end{figure}

\begin{figure}[t]
  \centering
  \includegraphics[width=\textwidth]{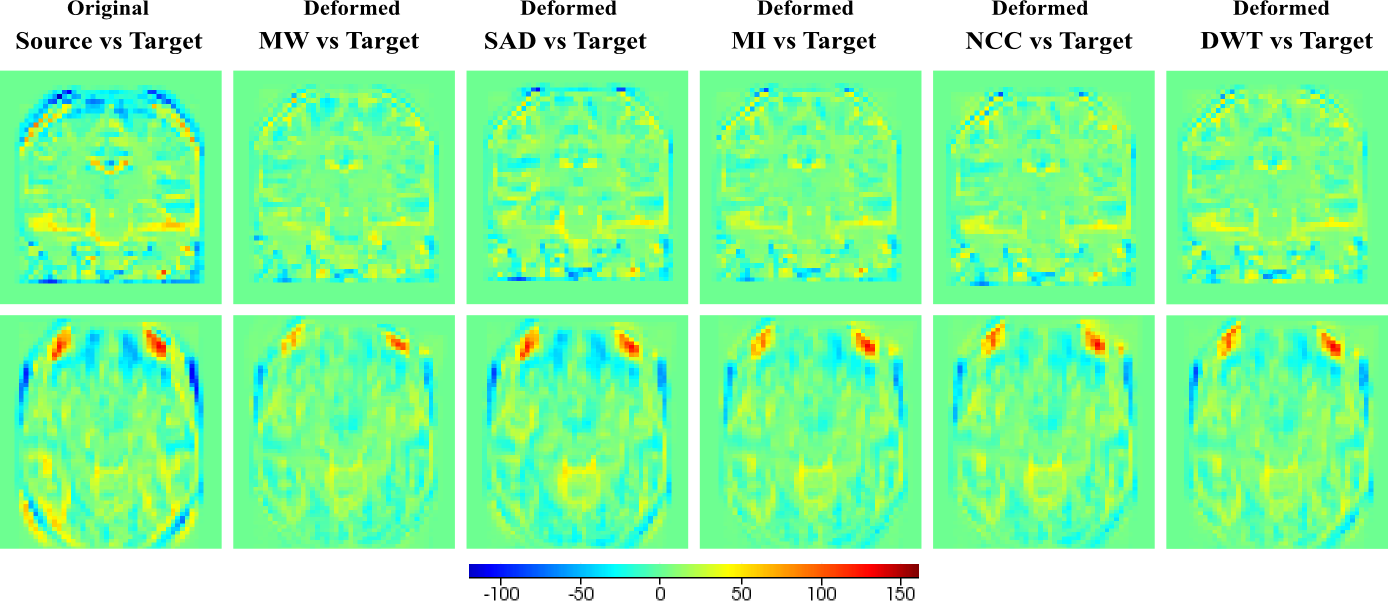}
  \caption{Qualitative results for one example slice from {\bf IBSR dataset}. In this case, since showing overlapped structures in the same image is too ambiguous given that the segmentation masks almost cover the complete image, we are showing the intensity difference between the two volumes. This is possible since images are coming from the same modality and they are normalized. The first column shows the difference of the original volumes before registration. From second to sixth column we observe the difference between the warped source and the target images, for the multiweight algorithm (MW) and the single metric algorithm using sum of absolute differences (SAD), mutual information (MI), normalized cross crorrelation (NCC) and discrete wavelet transform (DWT) as similarity measure. According to the scale in the bottom part of the image, extreme values (which mean high differences between the images) correspond to blue and red colors, while green indicates no difference in terms of intensity. Note how most of the big differences observed in the first column (before registration) are reduced in the multiweight algorithm, while some of them (specially in the peripheral area of the head) remain when using single metrics.} 
  \label{fig:resultsIBSR}
\end{figure}
\
\newpage


\afterpage{
\newpage
\section{Detailed version of the {\sc cccp} algorithm}
Here we include a detailed version of the {\sc cccp} algorithm presented as Algorithm 1 in the main paper.
\begin{algorithm}[h!]
\caption{The {\sc cccp} algorithm (detailed version).}
\label{algo:cccpDetailed}
\fontsize{11}{12}\selectfont
\begin{algorithmic}[1]
\State $\mathcal{D}$, ${\bf w}_0$, $C$, $\alpha$, $\eta$, the tolerance $\epsilon$.
\State $t \leftarrow 0$.
\State ${\bf w}_t \leftarrow {\bf w}_0$
\Repeat
\State For a given ${\bf w}_t$, impute the latent variables $\hat{D}_i$ for each sample by solving the problem: 
		\begin{eqnarray}
		\hat{D}_i = \argmin_{D \in \mathcal{L}^{|V|}} \Big( {\bf w}_t^\top \Psi({\bf x}_i, \Gamma) + \eta \Delta(g(S_i^I, D), S_i^J) \Big).  \nonumber
		\end{eqnarray}
		\State Initialize the constraint set for each sample: $\mathcal{W}_i \leftarrow \emptyset, \forall i$.
		\Repeat
		\State Obtain the most violated constraint (compute $\bar{D}_i$ for each sample):
		\begin{eqnarray}
		\bar{D}_i =  \argmin_{\bar{D} \in \mathcal{L}^{|V|}} \Big( {\bf w}^\top \Psi({\bf x}_i, \bar{D}) - \Delta(g(S_i^I, \bar{D}), S_i^J) \Big).   \nonumber
		\end{eqnarray}		 
		 \State Update constraint set if $\bar{D}_i$ is sufficiently violated.
		 \begin{equation}
		  \mathcal{W}_i \leftarrow \mathcal{W}_i \cup \bar{D}_i, \forall i. \nonumber
		 \end{equation}
		 
		\State Solve the following optimization problem to obtain ${\bf w}$:
		
		\begin{eqnarray}
			\label{eq:Obj}
			\min_{\bf w, \{\xi_i\}} && \frac{1}{2}||{\bf w}||^2 + \alpha ||{\bf w} - {\bf w}_0 ||^2 + \frac{C}{N} \sum_i \xi_i , \nonumber \\
s.t. && {\bf w}^\top \Psi({\bf x}_i, \hat{D}_i) \leq {\bf w}^\top \Psi({\bf x}_i, \bar{D_i}) - \Delta(g(S_i^I, \bar{D}), S_i^J) + \xi_i, \forall \bar{D_i} \in \mathcal{W}_i, \forall i \nonumber \\
&& w_p \geq 0, \xi_i \geq 0, \forall i.  \nonumber
		\end{eqnarray}
		\Until{No working set $\mathcal{W}_i$ can be further updated.}
		\State $t \leftarrow t+1$
		\State Update the parameters: ${\bf w}_t \leftarrow {\bf w}$
		\Until{Objective of the problem does not decrease more than $\epsilon$}.

    \end{algorithmic}
\end{algorithm}
}

\end{document}